%% file: sample_FG2023.tex
\documentclass[a4paper, 10pt, conference]{ieeeconf}      
\usepackage{bm}                                                         
\usepackage{FG2023}
\usepackage{amsmath}
\FGfinalcopy 
\usepackage{graphicx}
\usepackage{amsfonts}
\usepackage{algorithm}
\usepackage{physics}
\usepackage{algpseudocode}
\usepackage{booktabs}
\usepackage{subcaption}
\usepackage{cuted}
\usepackage{capt-of,etoolbox}
\usepackage{etoolbox,lipsum}
\usepackage{multirow}
\makeatletter
\newcommand{\algmargin}{\the\ALG@thistlm}
\makeatother
\newlength{\whilewidth}
\algdef{SE}[parWHILE]{parWhile}{EndparWhile}[1]
  {\parbox[t]{\dimexpr\linewidth-\algmargin}{%
     \hangindent\whilewidth\strut\algorithmicwhile\ #1\ \algorithmicdo\strut}}{\algorithmicend\ \algorithmicwhile}%
\algnewcommand{\parState}[1]{\State%
  \parbox[t]{\dimexpr\linewidth-\algmargin}{\strut #1\strut}}

\IEEEoverridecommandlockouts                              
\def\FGPaperID{20} 

\title{\LARGE \bf
T2V-DDPM: Thermal to Visible Face Translation using Denoising Diffusion Probabilistic Models
}

\author{\parbox{16cm}{\centering
    {\large Nithin Gopalakrishnan Nair and Vishal M. Patel}\\
    {\normalsize
    Department of Electrical and Computer Engineering, Johns Hopkins University, Baltimore, USA}}
    \thanks{This research was supported by NSF CAREER award 2045489.}
}
\begin{document}

\ifFGfinal
\thispagestyle{empty}
\pagestyle{empty}
\else
\author{Anonymous FG2023 submission\\ Paper ID \FGPaperID \\}
\pagestyle{plain}
\fi




\maketitle

\begin{abstract}
Modern-day surveillance systems perform person recognition using deep learning-based face verification networks. Most state-of-the-art facial verification systems are trained using visible spectrum images. But, acquiring images in the visible spectrum is impractical in scenarios of low-light and nighttime conditions, and often images are captured in an alternate domain such as the thermal infrared domain. Facial verification in thermal images is often performed after retrieving the corresponding visible domain images. This is a well-established problem often known as the Thermal-to-Visible (T2V) image translation.  In this paper, we propose a Denoising Diffusion Probabilistic Model (DDPM) based solution for T2V translation specifically for facial images. During training, the model learns the conditional distribution of visible facial images given their corresponding thermal image through the diffusion process. During inference, the visible domain image is obtained  by starting from Gaussian noise and performing denoising repeatedly. The existing inference process for DDPMs is stochastic and time-consuming. Hence, we propose a novel inference strategy for speeding up the inference time of DDPMs, specifically for the problem of T2V image translation. We achieve the state-of-the-art results on multiple datasets. The code and pretrained models are publically available at http://github.com/Nithin-GK/T2V-DDPM
\end{abstract}

\section{INTRODUCTION}
 
Many surveillance systems include sensors to capture images at multiple wavelengths to accommodate day and nighttime settings. Under low-light settings, the images captured by visible spectrum cameras fail to capture visual details in the scene due to the increasing amount of additive Poisson noise on the captured image. Hence, the use of infrared cameras to capture an additional thermal image is quite prominent. The images captured by these surveillance systems are further used for person recognition using facial recognition algorithms, and this falls under the broad area of research called Heterogeneous Face Recognition (HFR) \cite{li2007illumination,li2009hfb,liu2012heterogeneous}. Deep Convolutional Neural Network (CNN) based algorithms have produced state-of-the-art results for facial recognition with almost perfect accuracy in multiple benchmarks \cite{deng2019arcface,parkhi2015deep}. 
 
  Although the existing facial verification algorithms work remarkably well during daytime, the performance of these algorithms fall drastically under low-light settings, this is due to the large domain discrepancy between thermal and visible images. One solution is to retrain a network for facial recognition using just thermal images. But this is impractical since the normal CNN-based facial recognition networks \cite{deng2019arcface} often require large amounts of face images to obtain good performance and there does not exist many publicaly available large scale  thermal images to train such networks.  Hence the problem of HFR is very relevant and has multiple practical implications.
 \input{input_tex/noise_vis}
 
 The recent rise of conditional generative models \cite{goodfellow2014generative} has enabled an alternate approach to address this problem through a two-step process. First, the problem is simplified to an image-to-image translation problem from thermal to visible domains. After the corresponding visible images are obtained, facial recognition algorithms can be directly applied to them. Multiple works utilize GANs for tackling this problem \cite{mei2022escaping,di2019polarimetric,immidisetti2021simultaneous}. Conditional GANs attempt to learn the translation from the thermal domain to the visible domain through a min-max approach. Although with sufficient training, GAN-based T2V translation networks can produce good results, training of GANs is a tedious process and may result in phenomena like mode collapse. Also, given limited data, the convergence of GANs for accurate T2V translation is not guaranteed.
 
 Recently, Denoising Diffusion Probabilistic Models (DDPMs) \cite{ho2020denoising} have gained a significant attention due to their ability to generate high-quality images. Like Variational Autoencoders \cite{kingma2013auto}, DDPMs attempt to learn the variational lower bound of the log-likelihood of the data distribution. DDPMs have already beaten GANs in the task of image generation. Multiple methods have proposed DDPMs for low-level vision tasks like image super resolution, colorization, deblurring, and denoising \cite{saharia2021image,saharia2021palette,kawar2022denoising,lugmayr2022repaint}. These methods have performed much better in terms of quality of the output produced based on realness metrics like Fréchet Inception Distance (FID) \cite{heusel2017gans} and structural similarity metrics like Structural Similarity Index (SSIM). Moreover, there are works connecting the ability of DDPM to learn the conditional distribution to the optimal transport theory of the most efficient transformation \cite{su2022dual}. Despite its immense potential in modeling conditional distributions, there are no works yet published for the problem of T2V face image translation. In this paper, we propose a solution for T2V face translation using DDPMs.  Since the model attempts to learn the transformation, very few image pairs are required for training the model to achieve good results. We show this in Figure \ref{fig:facethvis} where we can clearly see the performance of DDPM-based models with just a few training image pairs. 
 
 One disadvantage of DDPMs over conventional CNNs is that the inference process is quite slow because of the underlying Markovian chain, which requires multiple forward passes through a single neural network. But through our experiments we have observed that the initial steps of the DDPM inference process aim to recover the high-level image details, like structure and color. This information can be easily captured from the corresponding thermal image; hence, rather than starting from isotropic Gaussian noise, we start with the noised thermal image. Thereby skipping some steps and speeding up the inference process.
 
 Infrared cameras capture images in different wavelengths within the infrared spectrum based on the application. Two common modalities are the Near Infrared (NIR) and Long wave Infrared (LWIR). The choice between NIR and LWIR is based on whether it is a short-range or long-range surveillance application. NIR images consist of images close to the visible spectrum; hence the resolution of these images is naturally high and can capture finer details in the facial image captured. LWIR images are often used for long-range surveillance applications, but these images are often shot in low-resolution and fail to capture essential facial details. This could be clearly seen when we compare the facial recognition accuracies when HFR algorithms are applied to these images. NIR face datasets \cite{duan2020cross,fu2019dual} produce accuracies close to $99\%$ but those of LWIR datasets \cite{immidisetti2021simultaneous,peri2021synthesis} are significantly lower. In this paper, we mainly focus on HFR for LWIR datasets.
 
 In short, this paper makes the following contributions
 \begin{itemize}
    \item We propose a DDPM-based solution for the HFR problem. We focus on the image translation problem from LWIR thermal images to their corresponding visible images. 
    \item We introduce a novel inference strategy for DDPMs mainly applicable to the T2V face translation problem. Through this we achieve upto $(2\times)$ speed up on the inference times without any drop in performance.
    \item We evaluate our method qualitatively and quantitatively on real-world datasets and show that it performs better than the existing methods on T2V face translation problem.
\end{itemize}

\begin{figure*}[t!]
	\centering
		\includegraphics[width=0.8\linewidth]{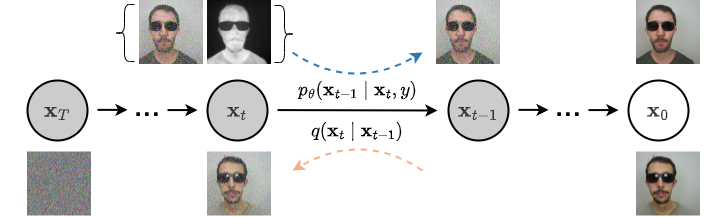}
	\centering
	\caption{An overview of the proposed method. During training the thermal image is conditioned along with the noisy visible image. During inference, we skip some timesteps by starting with a coarse image consisting of the background colour and the coarse features.}
	\label{fig:pipeline}
	\vskip 5pt
\end{figure*}
\section{Proposed Method}
\subsection{Denoising Diffusion Probabilistic Models}

Denoising Diffusion Probabilistic models  \cite{sohl2015deep,ho2020denoising} belong to a class of generative models where the models learn the distribution of data through a Markovian process. DDPMs consist of a forward process and a reverse process. The joint distribution of the data is represented by the reverse process.  The forward process is a Markovian process where the next state is obtained by sampling from a Gaussian distribution whose mean depends on the current state of the system, a predefined variance schedule as well as the time $t$.  The sampling operation for the forward step is given by
\begin{equation}
    q(x_t|x_{t-1}) = \mathcal{N}(x_t; \sqrt{1-\beta_t} x_{t-1}, \beta_t \bm{I}),
    \label{eq:q_sample}
\end{equation}
where $\{\beta_t\}$ is a predefined variance schedule. In practical scenarios, $t$ usually takes values between $10^{-4}$ to $10^{-2}$. This could also be considered as a noising operation, where the next state is obtained from the current state by adding a small Gaussian noise with variance schedule $\{\beta_t\}$. The state at timestep $t$ can also be computed from the initial state $x_0$. the distribution of the state at a particular timestep $t$ given $x_0$ is,
\begin{equation}
    q(x_t|x_{0}) = \mathcal{N}(x_t; \sqrt{\bar{\alpha}_t} x_0, (1-\bar{\alpha}_t) \bm{I}).
    \label{eq:q_sample0}
\end{equation}
Equivalently,
\begin{equation}
    x_t=\bar{\alpha}_t x_0+ (1-\bar{\alpha}_t) \epsilon, \epsilon \sim \mathcal{N}(0,I),
\end{equation}
where, $\bar{\alpha}_t=\prod_{s=1}^t\alpha_s$ and $\alpha_t=1-\beta_t$. As can be seen from Eq. \ref{eq:q_sample0}, for large $t$, $\bar{\alpha}_t$ becomes zero. Hence it results in a standard isotropic Gaussian. The reverse process denotes a generation step where we start from a standard Gaussian and iteratively perform denoising through $t$ timesteps to generate an image corresponding to the training distribution.  \cite{sohl2015deep} has shown that when the number of timesteps is large and the increment in $\{\beta_i\}$ is small, then the reverse distribution could be also approximated by a Gaussian. Each step in the reverse process is performed by sampling from a distribution whose parameters are modelled using a neural network with parameters $\theta$ and each reverse step is defined by,
\begin{equation}
    p_{\theta}(x_{t-1}|x_t) =\mathcal{N}(\mu_{\theta}(x_t,t),\Sigma_{\theta}(x_t,t)). 
\end{equation}
The parameters $\theta$ are obtained by minimizing the variational lower bound of the negative log-likelihood of the data distribution defined by
\begin{equation}
    E[-log(p(x_0)] \leq E_q[\frac{p_{\theta}(x_{0:T})}{q(x_{1:T}|x_0)}]=L.
\end{equation}
Further simplifying,
\begin{multline}
   L = \mathbb{E}_q \bigg[\underbrace{D_{KL}(q(\vb{x}_T|\vb{x}_0)\,||\,p(\vb{x}_T))}_{L_T} \underbrace{- \text{log}\,p_\theta(\vb{x}_0|\vb{x}_1)}_{L_0} \\
    + \sum_{t>1} \underbrace{D_{KL}(q(\vb{x}_{t-1}|\vb{x}_t, \vb{x}_0)\,||\,p(\vb{x}_{t-1}|\vb{x}_t))}_{L_{t-1}} \bigg].
\label{KL_eq}
\end{multline}

Recent works \cite{ho2020denoising} have shown the variance schedule for the reverse step could be kept the same as the forward process and only the means need to be learned using a neural network. Furthermore using the input at the current timestep, the mean of the reverse step could be reparametrized according to
\begin{equation}
    \mu_\theta (\vb{x}_t,t) = \frac{1}{\sqrt{\alpha_t}}\bigg( \vb{x}_t - \frac{\beta_t}{\sqrt{1-\Bar{\alpha}_t}}\epsilon_\theta(\vb{x}_t,t)\bigg).
    \label{mu eq}
    \end{equation}
Also given an $x_0$, $x_t$, the mean of the distribution $q(\vb{x}_{t-1}|\vb{x}_t, \vb{x}_0)$ can be written as 
\begin{equation}
    \mu (\vb{x}_t,t) = \frac{1}{\sqrt{\alpha_t}}\bigg( \vb{x}_t - \frac{\beta_t}{\sqrt{1-\Bar{\alpha}_t}}\epsilon(\vb{x}_t,t)\bigg).
    \label{mu eq}
    \end{equation}
Equating the means of the distributions, the training objective could be further\cite{ho2020denoising} simplified to,
\begin{equation}
    L_{simple} = \mathbb{E}_{t,\vb{x}_0,\epsilon} \big[ \|\epsilon - \epsilon_\theta(\vb{x}_t,t)\|^{2} \big].
\end{equation}
Once the data distribution is learned, new samples can be derived from the data distribution, starting from a Gaussian random sample and following a Markovian process.

\subsection{Conditional Diffusion Models for T2V Generation}
Saharia et al. \cite{saharia2021image} proposed a method for conditional generation using DDPM where the images are generated based on a constraint, and the model learns the conditional distribution given a condition $y$. Learning the conditional distribution rather than the unconditional distribution enables the use of DDPMs for low-level vision problems like image restoration. To make the model learn the conditional distribution, we condition the neural network with the thermal image $y$ at all timesteps $t$. Here the effective training objective is defined by,
\begin{equation}
    L_{simple} = \mathbb{E}_{t,\vb{x}_0,\epsilon} \big[ \|\epsilon - \epsilon_\theta(\vb{x}_t,y,t)\|^{2} \big].
\end{equation}
Once the conditional distribution is learnt, the model can be used for inference, i.e., given a thermal image, we can sample the corresponding visible image by starting with pure Gaussian noise and sampling from a Markov chain of $T$ steps. The corresponding mean of the conditional distribution for the transition $p_{\theta}(x_{t-1}|x_t,y)$ is
\begin{equation}
    \mu_\theta (\vb{x}_t,t) = \frac{1}{\sqrt{\alpha_t}}\bigg( \vb{x}_t - \frac{\beta_t}{\sqrt{1-\Bar{\alpha}_t}}\epsilon_\theta(\vb{x}_t,y,t)\bigg).
    \label{mu eq}
    \end{equation}
    and the variance schedule remains the same as that during training.
    
\subsection{Speeding up conditional diffusion models}
From our experiments we made a key observation that forms the basis for this section. As can be seen from Figure \ref{fig:noise_vis1}, which shows how the generated image varies with time, the low-level semantics of the image is learnt much further into the diffusion process. The initial steps are aimed at learning coarse features like the shape of the underlying face and background color. This means if we have an image with coarse facial features but lacks finer details, it can possibly be utilized to speed up the reconstruction process. Motivated by this empirical observation, we   generate an image with coarse features through the following process. Given the thermal image $y$ and the visible image, $x$ normalized in the range $[0,1]$, we binarize the thermal image to generate a mask $m$ through the following operation
\begin{equation}
    m= 
\begin{cases}
    1& \text{if } x > \epsilon\\
    0              & \text{otherwise}.
\end{cases}
\end{equation}
\input{input_tex/vis_timestep}

Since the background colour is constant for almost all images, we sample the background colour from a random image in the training dataset and form a grid $c$ of the size of the image. The coarse image $y^c$ is created according to 
\begin{equation}
    y^c = m\cdot y  +(1-m) \cdot c.
    \label{eq:coarse}
\end{equation}
Using $y^c$, the noisy version of $y^c$ denoted by $y^c_{T_r}$ corresponding to time $T_r$ is created. Figure \ref{fig:noise_vis1} shows a visualization of $y^c$, $y^c_{T_r}$ as well as $x_{T_r}$ starting from an isotropic Gaussian noise. From the visualization, one can note that $x_{T_r}$ and $y^c_{T_r}$ for $T_r=50$ does not differ much semantically. Moreover the SNR values of both the images are almost the same.

\begin{algorithm}
\caption{Training}
\begin{algorithmic}[1]
\renewcommand{\algorithmicrequire}{\textbf{Input:}}
\renewcommand{\algorithmicensure}{\textbf{Initialize:}}
\Require Thermal image and visible image pairs $P = \{(\vb{y}^k,\vb{x}^k)\}^{K}_{k=1}$
\Repeat
    \State $(\vb{y},\vb{x})\sim P$
    \State $t \sim \text{Uniform}(\{1,\ldots,T\})$
    \State $\epsilon\sim \mathcal{N}(\vb{0}, \vb{I})$
    \State  $\vb{x}_t = \sqrt{\bar{\alpha_t}\vb{x}_c + \sqrt{1-\bar{\alpha_t}}\epsilon}$
    \State Gradient descent step on  $\nabla_\theta \,||\epsilon - \epsilon_\theta(\vb{x}_t,\vb{y}, t)||^{2}$
\Until {converged}
\end{algorithmic} 
\vspace{-1mm}
\label{train algo}
\end{algorithm}

\begin{algorithm}
\caption{Inference}
\begin{algorithmic}[1]
\renewcommand{\algorithmicrequire}{\textbf{Input:}}
\renewcommand{\algorithmicensure}{\textbf{Initialize:}}
\Require Thermal image $\vb{x}$, background colour $c$ 
\State Create course image $x^c$ according to Equation \ref{eq:coarse}
\State Sample $x^c_{T_r}\sim q(x^c_{T_r}|x^c)$
\For{$t = T_r, \ldots, 1$}
    \State sample $\vb{z}\sim \mathcal{N}(\vb{0},\vb{I})$ if $t>1$ else $\vb{z}=0$
    \parState{ %
compute $\vb{x}^c_{t-1} = \frac{1}{\sqrt{\alpha_t}}\Big(\vb{x}^c_t - \frac{1-\alpha_t}{\sqrt{1-\bar{\alpha_t}}} \epsilon_\theta(\vb{x}^c_t,\vb{x},t)\Big) + \sigma_t\vb{z}$}
\EndFor

\\
\Return $x^c_0$
\end{algorithmic}
\label{test algo}
\end{algorithm}
Algorithm \ref{train algo} summarises the training procedure of the proposed method. The thermal image and their corresponding noisy visible images are used for training a DDPM model for $T$ timesteps. During inference, we reduce the inference time of the diffusion process by starting from a noisy coarse image than isotropic Gaussian noise. The inference procedure is given in Algorithm \ref{test algo}.
\input{input_tex/THVIS_results}

\section{Experiments}
In our experiments, we focus on the translation problem from thermal images of dimension $128 \times 128$ to visible images of size $128\times 128$. The main emphasis of our experiments is on improving the facial recognition accuracy of the reconstructed images, but we also perform comparisons with different metrics. Until now, there are no standardized baselines for T2V face translation. Hence we make use of the datasets used in the work \cite{mei2022escaping} for our experiments. We perform experiments on three different datasets for T2V face translation.  More details about the individual datasets and their evaluation criteria are presented in this section.

\noindent\textbf{VIS-TH dataset \cite{mei2022escaping}:-} The VIS-TH dataset consists of facial images corresponding to 50 different identities. Images in the VIS-TH dataset are captured through a dual-sensor camera in Long Wave Infrared (LWIR) modality and are aligned. The facial images in the dataset consist of various poses. We create the train-test split for the dataset in the same fashion as followed by \cite{mei2022escaping,immidisetti2021simultaneous}. We use all the images corresponding to 40 randomly selected identities as the training set and the remaining images as the test set. VIS-TH is quite a challenging dataset owing to the low number of identities and the diversity of the dataset.

\noindent\textbf{ARL-VTF dataset\cite{poster2021large}:-} Like the VIS-TH dataset, the RL-VTF dataset also consists of facial images captured in the LWIR modality. The dataset also provides the image capture settings for aligning the faces. Visible images in the ARL-VTF dataset are severely overexposed. Hence we correct this overexposure through exposure matching with the VIS-TH dataset. We create a subset of the original ARL-VTF dataset for all our experiments and choose 100 identities with different expressions as the training dataset, and data corresponding to 40 identities as the testing set. In total, there are 3,200 training pairs and 985 testing pairs.

\noindent\textbf{Evaluation metrics:} For evaluating the effectiveness of our method, we utilize two different schemes like \cite{duan2020cross,mei2022escaping}. We evaluate the facial verification performance of the reconstructed images by using and comparing our method with the existing methods in terms of the Rank-1 accuracy, Verification Rate (VR) @ False Accept Rate (FAR)=$1\%$ and
VR@FAR=$0.1\%$. All Facial verification experiments are conducted using pretrained ArcFace facial recongittion system\cite{deng2019arcface}. For evaluating the quality of the reconstructed outputs, we use the following metrics: Learned Perceptual Image Patch Similarity (LPIPS) \cite{zhang2018perceptual}, Deg (cosine distance between LightCNN  features), Peak Signal to Noise ratio (PSNR) of the underlying grayscale image and Structural Similarity Index (SSIM).

\noindent\textbf{Training settings:} For the diffusion model, we use the same parameters as for the iamgeNet superresolution model as used in improved diffusion \cite{nichol2021improved}. We initialize the model with the imageNet pretrained weights. The model is trained for $T=1000$ timesteps. During inference, we use timestep rescaling as in \cite{nichol2021improved} and reduce the number of inference steps to $100$. $T_r=60$, $\epsilon$ is set equal to $0.1$ for all experiments

\noindent\textbf{Comparison methods:} We evaluate our method by comparing with different generative model based approaches for image-to-image translation. The following methods are used for comparison: Pixel2Pixel\cite{isola2017image}, Self-Attention GAN (SAGAN)\cite{di2019polarimetric}, GANVFS\cite{zhang2019synthesis}, HIFaceGAN\cite{yang2020hifacegan} and AxialGAN\cite{immidisetti2021simultaneous}.

\input{input_tex/ARL_VTF_results}


\begin{table}[tp!]
	\caption{Verification results on VIS-TH dataset}
	\begin{center}
		\resizebox{1\linewidth}{!}{
			\begin{tabular}{ c c c c }
				\hline
				Method &Rank-1&VR@FAR=1\%   &VR@FAR=0.1\% \\
				\hline\
				Pix2Pix\cite{isola2017image}&20.88&1.73&0.0   \\
				SAGAN\cite{di2019polarimetric}&13.04&0.0&0.0  \\
				GANVFS\cite{zhang2019synthesis}& 30.98&3.43&0.0\\
				HiFaceGAN\cite{yang2020hifacegan}& 68.48&37.14&17.14   \\
				AxialGAN\cite{immidisetti2021simultaneous}& 58.15&23.43&6.86  \\
				\hline
					T2V-DDPM& 70.5&45.5&23.5   \\
					\hline
			\end{tabular}
		}
	\end{center}
	\label{table:visth_ver}
	\vspace{-5mm}
\end{table}
\begin{table}[htp!]
	\caption{Image Quality metrics results on VIS-TH dataset}
	\begin{center}
		\resizebox{1\linewidth}{!}{
			\begin{tabular}{ c| c| c|c c }
				\hline
				Method &LPIPS($\downarrow$)&Deg.($\uparrow$)&PSNR($\uparrow$)&SSIM($\uparrow$) \\
				\hline
				TH&0.5780&0.3635&6.3516&0.1841\\
				\hline
				Pix2Pix\cite{isola2017image}&0.4847&0.2794&13.45&0.353   \\
				SAGAN\cite{di2019polarimetric}&0.4678&0.3367&14.14&0.4183 \\
				GANVFS\cite{zhang2019synthesis}& 0.4116&0.3292&13.83&0.4186\\
				HiFaceGAN\cite{yang2020hifacegan}& 0.1950&0.6268&19.16&0.6963  \\
				AxialGAN\cite{immidisetti2021simultaneous}&0.1756&0.6157&22.80&0.7458  \\
				\hline
					T2V-DDPM& 0.1718&0.6441&19.419&0.7178  \\
					\hline
			\end{tabular}
		}
	\end{center}
	\label{table:visth_img}
	\vspace{-5mm}
\end{table}
\subsection{Results on the VIS-TH dataset}
Figure \ref{fig:facethvis} shows the qualitative results on the VIS-TH dataset for four different poses. As we can see, all the existing methods except HIFaceGAN fail to reconstruct attributes like hats and sunglasses. Whereas our method, as well as HIFaceGAN work well even with these occlusions. If we have a close look at the last row of Figure \ref{fig:facethvis} we can see that when the frontal pose changes, HiFaceGAN fails to reconstruct the face properly, but our method works well even with different attributes as well as poses. For evaluating the methods quantitatively, we use two criteria as mentioned earlier. The quantitative results can be found in Table \ref{table:visth_img} and \ref{table:visth_ver}. For the rank-1 accuracy, we can see that we gain an improvement by $2\% $. For VR@FAR=$0.1\%$ and VR@FAR=$1\%$, we gain performance improvements of $8.3\%$ and $6.3\%$ respectively. For the image quality metrics, in terms of the perceptual similarity metrics (LPIPS and Deg), we gain an improvement of $0.0038$ and $0.03$, respectively. Our values for PSNR and SSIM are a bit lower than  AxialGAN.  However, note that PSNR often does not often represent the amount of facial details present in the reconstructed images since a blurry image could have a higher PSNR compared to a much sharper image.

\subsection{Results on the ARL-VTF dataset}
Figure \ref{fig:facearlvtf} shows the qualitative results on the ARL-VTF dataset. This dataset consists of more identities and much more training and testing images. As we can see in Fig. \ref{fig:facearlvtf},  only our method can properly reconstruct the salient underlying facial features, and it can create realistic features that have coarse features corresponding to the thermal image. Further quantitative evaluations can be found in Table \ref{table:arlvtf_img} and \ref{table:arlvtf_ver}.
\begin{table}[tp!]
	\caption{Verification results on the ARL-VTF dataset}
	\begin{center}
		\resizebox{1\linewidth}{!}{
			\begin{tabular}{ c c c c }
				\hline
				Method &Rank-1&VR@FAR=1\%   &VR@FAR=0.1\% \\
				\hline
				Pix2Pix\cite{isola2017image}&18.88&5.09&0.33   \\
				SAGAN\cite{di2019polarimetric}&13.46&5.42&0.33 \\
				GANVFS\cite{zhang2019synthesis}& 21.84&12.97&2.63\\
				HiFaceGAN\cite{yang2020hifacegan}& 65.35&41.33&20.89  \\
				AxialGAN\cite{immidisetti2021simultaneous}& 66.67&42.86&18.62 \\
				\hline
					T2V-DDPM& 75.37&43.51&19.87  \\
					\hline
			\end{tabular}
		}
	\end{center}
	\label{table:arlvtf_ver}
	\vspace{-5mm}
\end{table}
\begin{table}[tp!]
	\caption{Image quality metrics results on the ARL-VTF dataset}
	\begin{center}
		\resizebox{1\linewidth}{!}{
			\begin{tabular}{ c|  c| c|c c }
				\hline
				Method &LPIPS($\downarrow$)&Deg.($\uparrow$)&PSNR($\uparrow$)&SSIM($\uparrow$) \\
				\hline
				TH&0.5551&0.4201&5.674&0.1095\\
				\hline
				Pix2Pix\cite{isola2017image}&0.4467&0.3489&13.12&0.3804  \\
				SAGAN\cite{di2019polarimetric}&0.4044&0.4244&14.25&0.4490 \\
				GANVFS\cite{zhang2019synthesis}& 0.3924&0.3935&13.68&0.4313\\
				HiFaceGAN\cite{yang2020hifacegan}& 0.1937&0.6563&19.62&0.6937 \\
				AxialGAN\cite{immidisetti2021simultaneous}&0.2123&0.6699&20.04&0.7179
				\\
				\hline
					T2V-DDPM& 0.2010&0.6606&19.70&0.6775 \\
					\hline
			\end{tabular}
		}
	\end{center}
	\label{table:arlvtf_img}
	\vspace{-5mm}
\end{table}
\subsection{Ablation study: Performance variation on changing $T_r$:}
To note the performance variation on changing $T_r$, we vary $T_r$ in steps of 10 from $T_r=40,...,100$  on the VIS-TH dataset, where $T_r=100$ denotes starting from pure Gaussian noise. The corresponding results are shown in Table \ref{table:ablation}. As can be seen,  the facial verification performance doesn't vary much even for $T_r=40$. This clearly validates our claim and observation that in the initial steps of the diffusion process, only the coarse features are learnt and if this could be inputted properly to the diffusion model, we can achieve major boost for the inference time.
\begin{table}[tp!]
	\caption{Ablation study on the TH-VIS dataset. The number of inference timesteps are varied here and the performance is noted.}
	\begin{center}
		\resizebox{1\linewidth}{!}{
			\begin{tabular}{ c |c c c }
				\hline
				$T_r$ &Rank-1&VR@FAR=1\%   &VR@FAR=0.1\% \\
				\hline
				40&66.5&34.0&18.0\\
				50&70.5&45.5&23.5   \\
				60&71&42.5&31.0 \\
				70&70.5&40.0&21.0\\
				80&68.5&39.0&23.5   \\
				90&72.0&37.0&22.0 \\
				100& 70.0&40.5&18.5  \\
					\hline
			\end{tabular}
		}
	\end{center}
	\label{table:ablation}
	\vspace{-5mm}
\end{table}

\section{Conclusion}
We presented a solution for V2T face translation using DDPMs by treating it as a conditional image generation problem. We find that with the immense power in modeling probabilistic distributions, DDPMs prove as an ideal solution for generating samples from the conditional distribution of visible images given thermal images. We also introduce a novel sampling strategy to reduce the inference time for DDPMs. Our experiments on multiple datasets show that DDPMs perform better than GANs for the T2V face translation problem. To the best of our knowledge, this is the first work utilizing DDPMs to reconstruct visible facial images from thermal images.
\section{ACKNOWLEDGMENTS}

This research was supported by NSF CAREER award 2045489.


{\small
\bibliographystyle{ieee}
\bibliography{egbib}
}

\end{document}

%% file: input_tex/noise_vis.tex
\begin{figure*}[!htb]
 \centering
    \begin{subfigure}[t]{0.115\linewidth}
      \captionsetup{justification=centering, labelformat=empty, font=scriptsize}
    \includegraphics[width=1\linewidth]{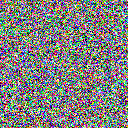}
     \includegraphics[width=1\linewidth]{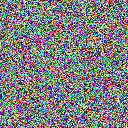}
      \caption{$T=100$}
    \end{subfigure}
    \begin{subfigure}[t]{0.115\linewidth}
      \captionsetup{justification=centering, labelformat=empty, font=scriptsize}
    \includegraphics[width=1\linewidth]{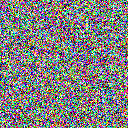}
     \includegraphics[width=1\linewidth]{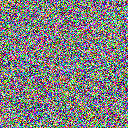}
      \caption{$T=86$}
    \end{subfigure}
    \begin{subfigure}[t]{0.115\linewidth}
      \captionsetup{justification=centering, labelformat=empty, font=scriptsize}
    \includegraphics[width=1\linewidth]{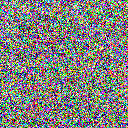}
     \includegraphics[width=1\linewidth]{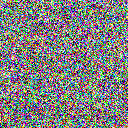}
      \caption{$T=72$}
    \end{subfigure}
    \begin{subfigure}[t]{0.115\linewidth}
      \captionsetup{justification=centering, labelformat=empty, font=scriptsize}
    \includegraphics[width=1\linewidth]{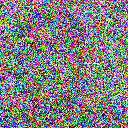}
     \includegraphics[width=1\linewidth]{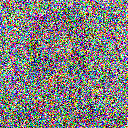}
      \caption{$T=58$}
    \end{subfigure}
    \begin{subfigure}[t]{0.115\linewidth}
      \captionsetup{justification=centering, labelformat=empty, font=scriptsize}
    \includegraphics[width=1\linewidth]{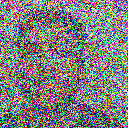}
     \includegraphics[width=1\linewidth]{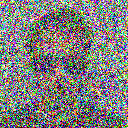}
      \caption{$T=44$}
    \end{subfigure}
    \begin{subfigure}[t]{0.115\linewidth}
      \captionsetup{justification=centering, labelformat=empty, font=scriptsize}
    \includegraphics[width=1\linewidth]{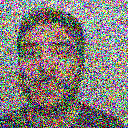}
     \includegraphics[width=1\linewidth]{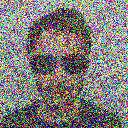}
      \caption{$T=29$}
    \end{subfigure}
    \begin{subfigure}[t]{0.115\linewidth}
      \captionsetup{justification=centering, labelformat=empty, font=scriptsize}
    \includegraphics[width=1\linewidth]{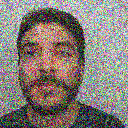}
     \includegraphics[width=1\linewidth]{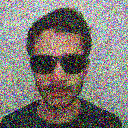}
      \caption{$T=15$}
    \end{subfigure}
 \begin{subfigure}[t]{0.115\linewidth}
      \captionsetup{justification=centering, labelformat=empty, font=scriptsize}
    \includegraphics[width=1\linewidth]{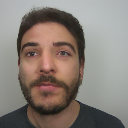}
     \includegraphics[width=1\linewidth]{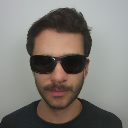}
    \caption{$T=0$}
    \end{subfigure}
    \caption{Visualization of time re-scaled diffusion process during inference.}
    \label{fig:noise_vis}
  \end{figure*}

%% file: input_tex/vis_timestep.tex
\begin{figure*}[!htb]
 \centering
    \begin{subfigure}[t]{0.13\linewidth}
      \captionsetup{justification=centering, labelformat=empty, font=scriptsize}
    \includegraphics[width=1\linewidth]{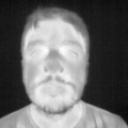}
     \includegraphics[width=1\linewidth]{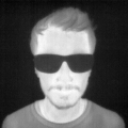}
      \caption{$y$}
    \end{subfigure}
    \begin{subfigure}[t]{0.13\linewidth}
      \captionsetup{justification=centering, labelformat=empty, font=scriptsize}
    \includegraphics[width=1\linewidth]{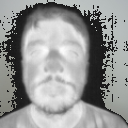}
     \includegraphics[width=1\linewidth]{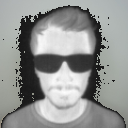}
      \caption{$y^c$}
    \end{subfigure}
    \begin{subfigure}[t]{0.13\linewidth}
      \captionsetup{justification=centering, labelformat=empty, font=scriptsize}
    \includegraphics[width=1\linewidth]{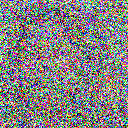}
     \includegraphics[width=1\linewidth]{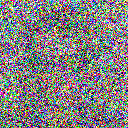}
      \caption{$y^c_{t1}$}
    \end{subfigure}
    \begin{subfigure}[t]{0.13\linewidth}
      \captionsetup{justification=centering, labelformat=empty, font=scriptsize}
    \includegraphics[width=1\linewidth]{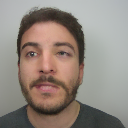}
     \includegraphics[width=1\linewidth]{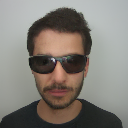}
      \caption{$x_0^c$}
    \end{subfigure}
  \begin{subfigure}[t]{0.13\linewidth}
      \captionsetup{justification=centering, labelformat=empty, font=scriptsize}
    \includegraphics[width=1\linewidth]{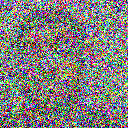}
     \includegraphics[width=1\linewidth]{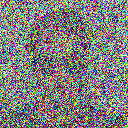}
      \caption{$x^c_{50}$}
    \end{subfigure}
    \begin{subfigure}[t]{0.13\linewidth}
      \captionsetup{justification=centering, labelformat=empty, font=scriptsize}
    \includegraphics[width=1\linewidth]{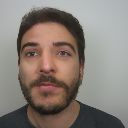}
     \includegraphics[width=1\linewidth]{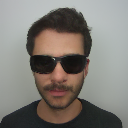}
      \caption{$x^c_0$}
    \end{subfigure}
    \begin{subfigure}[t]{0.13\linewidth}
      \captionsetup{justification=centering, labelformat=empty, font=scriptsize}
    \includegraphics[width=1\linewidth]{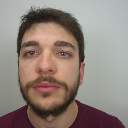}
     \includegraphics[width=1\linewidth]{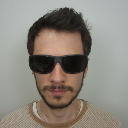}
      \caption{$x$}
    \end{subfigure}
    \caption{(a) Thermal Image, (b) Masked thermal image (c) Noise masked thermal image(T=50) (d) Reconstructed masked thermal image(T=100) (e) Intermediate diffusion process result image starting from Gaussian noise visualized at (T=50) (f) Reconstructed image starting from (T=0). (g) Visible Image }
        \label{fig:noise_vis1}
  \end{figure*}

%% file: input_tex/THVIS_results.tex
\begin{figure*}[!htb]
 \centering
    \begin{subfigure}[t]{0.115\linewidth}
      \captionsetup{justification=centering, labelformat=empty, font=scriptsize}
    \includegraphics[width=1\linewidth]{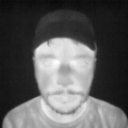}
    \includegraphics[width=1\linewidth]{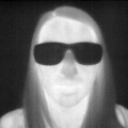}
    \includegraphics[width=1\linewidth]{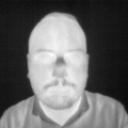}
    \includegraphics[width=1\linewidth]{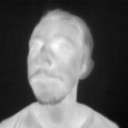}

      \caption{THERMAL}
    \end{subfigure}
    \begin{subfigure}[t]{0.115\linewidth}
      \captionsetup{justification=centering, labelformat=empty, font=scriptsize}
    \includegraphics[width=1\linewidth]{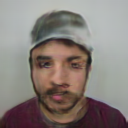}
    \includegraphics[width=1\linewidth]{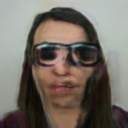}
    \includegraphics[width=1\linewidth]{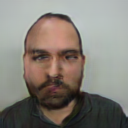}
    \includegraphics[width=1\linewidth]{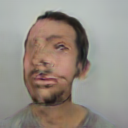}
      \caption{PIX2PIX\cite{isola2017image}}
    \end{subfigure}
    \begin{subfigure}[t]{0.115\linewidth}
      \captionsetup{justification=centering, labelformat=empty, font=scriptsize}
    \includegraphics[width=1\linewidth]{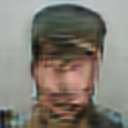}
    \includegraphics[width=1\linewidth]{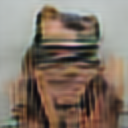}
    \includegraphics[width=1\linewidth]{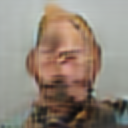}
    \includegraphics[width=1\linewidth]{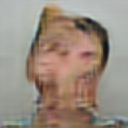}
      \caption{SAGAN\cite{di2019polarimetric}}
    \end{subfigure}
    \begin{subfigure}[t]{0.115\linewidth}
      \captionsetup{justification=centering, labelformat=empty, font=scriptsize}
    \includegraphics[width=1\linewidth]{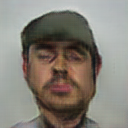}
    \includegraphics[width=1\linewidth]{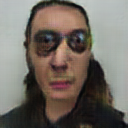}
    \includegraphics[width=1\linewidth]{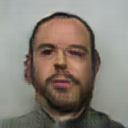}
    \includegraphics[width=1\linewidth]{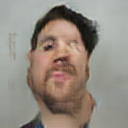}
      \caption{GANVFS\cite{zhang2019synthesis}}
    \end{subfigure}
    \begin{subfigure}[t]{0.115\linewidth}
      \captionsetup{justification=centering, labelformat=empty, font=scriptsize}
    \includegraphics[width=1\linewidth]{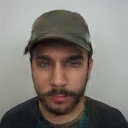}
    \includegraphics[width=1\linewidth]{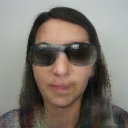}
    \includegraphics[width=1\linewidth]{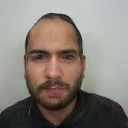}
    \includegraphics[width=1\linewidth]{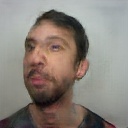}
      \caption{HIFACEGAN\cite{yang2020hifacegan}}
    \end{subfigure}
    \begin{subfigure}[t]{0.115\linewidth}
      \captionsetup{justification=centering, labelformat=empty, font=scriptsize}
    \includegraphics[width=1\linewidth]{compiled_single/axialgan_mod/Epoch49_046_1_14_OH.png}
    \includegraphics[width=1\linewidth]{compiled_single/axialgan_mod/Epoch50_004_1_13_OSG.png}
    \includegraphics[width=1\linewidth]{compiled_single/axialgan_mod/Epoch50_036_1_03_EA.png}
    \includegraphics[width=1\linewidth]{compiled_single/axialgan_mod/Epoch50_043_1_11_PR.png}
      \caption{AXIALGAN\cite{immidisetti2021simultaneous}}
    \end{subfigure}
    \begin{subfigure}[t]{0.115\linewidth}
      \captionsetup{justification=centering, labelformat=empty, font=scriptsize}
    \includegraphics[width=1\linewidth]{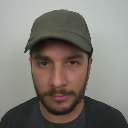}
    \includegraphics[width=1\linewidth]{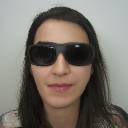}
    \includegraphics[width=1\linewidth]{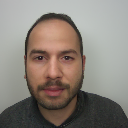}
    \includegraphics[width=1\linewidth]{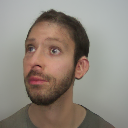}
      \caption{OURS}
    \end{subfigure}
 \begin{subfigure}[t]{0.115\linewidth}
      \captionsetup{justification=centering, labelformat=empty, font=scriptsize}
    \includegraphics[width=1\linewidth]{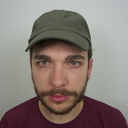}
    \includegraphics[width=1\linewidth]{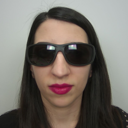}
    \includegraphics[width=1\linewidth]{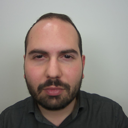}
    \includegraphics[width=1\linewidth]{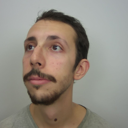}
    \caption{VISIBLE}
    \end{subfigure}
    \caption{Qualitative results for the thermal to visible translation tasks for facial images on a challenging dataset (VIS-TH dataset). Due to the low number of training image pairs, existing methods fail to reconstruct facial images accuractely, see coloumns(1,2,3,4) in the results and doesn't generalize well over unseen poses. On contrast Our method is able to generalize well given a low number of training image pairs.}
    \label{fig:facethvis}
  \end{figure*}

%% file: input_tex/ARL_VTF_results.tex
\begin{figure*}[!htb]
 \centering
    \begin{subfigure}[t]{0.115\linewidth}
      \captionsetup{justification=centering, labelformat=empty, font=scriptsize}
    \includegraphics[width=1\linewidth]{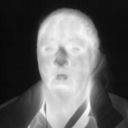}
    \includegraphics[width=1\linewidth]{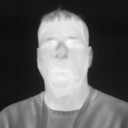}
    \includegraphics[width=1\linewidth]{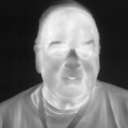}
      \caption{THERMAL}
    \end{subfigure}
    \begin{subfigure}[t]{0.115\linewidth}
      \captionsetup{justification=centering, labelformat=empty, font=scriptsize}
    \includegraphics[width=1\linewidth]{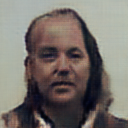}
    \includegraphics[width=1\linewidth]{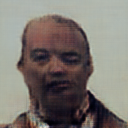}
    \includegraphics[width=1\linewidth]{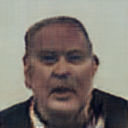}
      \caption{PIX2PIX\cite{isola2017image}}
    \end{subfigure}
    \begin{subfigure}[t]{0.115\linewidth}
      \captionsetup{justification=centering, labelformat=empty, font=scriptsize}
    \includegraphics[width=1\linewidth]{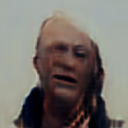}
    \includegraphics[width=1\linewidth]{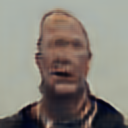}
    \includegraphics[width=1\linewidth]{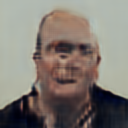}
      \caption{SAGAN\cite{di2019polarimetric}}
    \end{subfigure}
    \begin{subfigure}[t]{0.115\linewidth}
      \captionsetup{justification=centering, labelformat=empty, font=scriptsize}
    \includegraphics[width=1\linewidth]{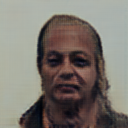}
    \includegraphics[width=1\linewidth]{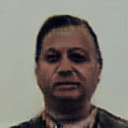}
    \includegraphics[width=1\linewidth]{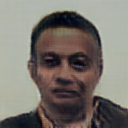}
      \caption{GANVFS\cite{di2019polarimetric}}
    \end{subfigure}
    \begin{subfigure}[t]{0.115\linewidth}
      \captionsetup{justification=centering, labelformat=empty, font=scriptsize}
    \includegraphics[width=1\linewidth]{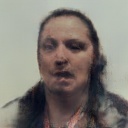}
    \includegraphics[width=1\linewidth]{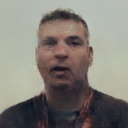}
    \includegraphics[width=1\linewidth]{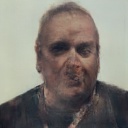}
      \caption{HIFACEGAN\cite{yang2020hifacegan}}
    \end{subfigure}
    \begin{subfigure}[t]{0.115\linewidth}
      \captionsetup{justification=centering, labelformat=empty, font=scriptsize}
    \includegraphics[width=1\linewidth]{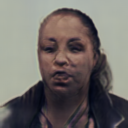}
    \includegraphics[width=1\linewidth]{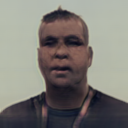}
    \includegraphics[width=1\linewidth]{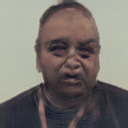}
      \caption{AXIALGAN\cite{immidisetti2021simultaneous}}
    \end{subfigure}
    \begin{subfigure}[t]{0.115\linewidth}
      \captionsetup{justification=centering, labelformat=empty, font=scriptsize}
    \includegraphics[width=1\linewidth]{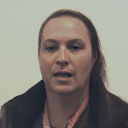}
    \includegraphics[width=1\linewidth]{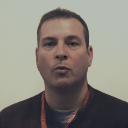}
    \includegraphics[width=1\linewidth]{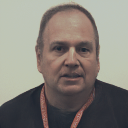}
      \caption{OURS}
    \end{subfigure}
 \begin{subfigure}[t]{0.115\linewidth}
      \captionsetup{justification=centering, labelformat=empty, font=scriptsize}
    \includegraphics[width=1\linewidth]{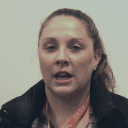}
    \includegraphics[width=1\linewidth]{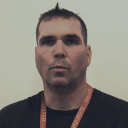}
    \includegraphics[width=1\linewidth]{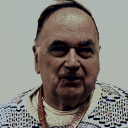}
    \caption{VISIBLE}
    \end{subfigure}
    \caption{Qualitative results on the ARL-VTF-dataset.}
    \label{fig:facearlvtf}
  \end{figure*}